\documentclass[11pt,a4paper]{article}
\usepackage[hyperref]{emnlp-ijcnlp-2019}
\usepackage{times}
\usepackage{latexsym}
\usepackage{csquotes}

\usepackage{url}

\aclfinalcopy 

\usepackage{booktabs} 
\usepackage{latexsym}
\usepackage{url}
\usepackage{enumerate}
\usepackage{graphicx}
\usepackage{epstopdf}

\usepackage{tabularx}
\usepackage{multirow}
\usepackage{subcaption}
\usepackage{tabularx}

\title{BillSum: A Corpus for Automatic Summarization of US Legislation}

\author{Anastassia Kornilova \\
  FiscalNote Research\\
  Washington, DC \\
  {\tt anastassia@fiscalnote.com} \\\And
  Vlad Eidelman \\
  FiscalNote Research\\
  Washington, DC \\
  {\tt vlad@fiscalnote.com} \\}

\date{}

\begin{document}
\maketitle

\begin{abstract}

Automatic summarization methods have been studied on a variety of domains, including news and scientific articles. Yet, legislation has not previously been considered for this task, despite US Congress and state governments releasing tens of thousands of bills every year. In this paper, we introduce BillSum, the first dataset for summarization of US Congressional and California state bills (\url{https://github.com/FiscalNote/BillSum}). We explain the properties of the dataset that make it more challenging to process than other domains. Then, we benchmark extractive methods that consider neural sentence representations and traditional contextual features. Finally, we demonstrate that models built on Congressional bills can be used to summarize California bills, thus, showing that methods developed on this dataset can transfer to states without human-written summaries.

\end{abstract}

\section{Introduction}

The growing number of publicly available documents produced in the legal domain has led political scientists, legal scholars, politicians, lawyers, and citizens alike to increasingly adopt computational tools to discover and digest relevant information. In the US Congress, over 10,000 bills are introduced each year, with state legislatures introducing tens of thousands of additional bills. Individuals need to quickly process them, but these documents are often long and technical, making it difficult to identify the key details.
While each US bill comes with a human-written summary from the Congressional Research Service (CRS),\footnote{\url{http://www.loc.gov/crsinfo/}} similar summaries are not available in most state and local legislatures. 

Automatic summarization methods aim to condense an input document into a shorter text while retaining the salient information of the original. 
To encourage research into automatic legislative summarization, we introduce the BillSum dataset, which contains a primary corpus of 22,218 US Congressional bills and reference summaries split into a train and a test set. Since the motivation for this task is to apply models to new legislatures, the corpus contains an additional test set of 1,237 California bills and reference summaries. We establish several benchmarks and show that there is ample room for new methods that are better suited to summarize technical legislative language.




\section{Background}

Research into automatic summarization has been conducted in a variety of domains, such as news articles \cite{cnndailymail}, emails \cite{nenkova2004email}, scientific papers \cite{teufelscience, collins2017supervised}, and court proceedings \cite{grover-etal-2004-holj, rhetorical_roles, kim2012summarization}. The later area is most similar to BillSum in terms of subject matter. However, the studies in that area either apply traditional domain-agnostic techniques or take advantage of the unique structures that are consistently present in legal proceedings (e.g precedent, law, background).\footnote{\newcite{kanapala2017text} provide a comprehensive overview of the works in legal summarization.}


While automatic summarization methods have not been applied to legislative text, previous works have used the text to automatically predict bill passage and legislators' voting behavior \cite{gerrish2011predicting, yano2012textual, eidelman2018predictable, kornilova2018party}. However, these studies treated the document as a ``bag-of-words'' and did not consider the importance of individual sentences. Recently, documents from state governments have been subject to syntactic parsing for knowledge graph construction \cite{kalouli2018cousbi} and textual similarity analysis~\cite{text_reuse}. Yet, to the best of our knowledge, BillSum is the first corpus designed, specifically for summarization of legislation.



\section{Data}
\label{sec:data}
The BillSum dataset consists of three parts: US training bills, US test bills and California test bills. The US bills were collected from the \textbf{Govinfo} service provided by the United States Government Publishing Office (GPO).\footnote{\url{https://github.com/unitedstates/congress}} Our corpus consists of bills from the 103rd-115th (1993-2018) sessions of Congress. The data was split into 18,949 train bills and 3,269 test bills. For California, bills from the 2015-2016 session were scraped directly from the legislature's website;\footnote{\url{http://leginfo.legislature.ca.gov}} the summaries were written by their Legislative Counsel.

The BillSum corpus focuses on mid-length legislation from 5,000 to 20,000 character in length. We chose to measure the text length in characters, instead of words or sentences, because the texts have complex structure that makes it difficult to consistently measure words. The range was chosen because on one side, short bills introduce minor changes and do not require summaries. While the CRS produces summaries for them, they often contain most of the text of the bill. On the other side, very long legislation is often composed of several large sections. The summarization problem thus becomes more akin in its formulation to multi-document summarization, a more challenging task that we leave to future work. The resulting corpus includes about 20\% of all US bills from this time period, where a majority of removed bills are either shorter than 5000 characters or identified as a near duplicate of a bill in the dataset.\footnote{Often the same bill is introduced multiple times, either across chambers or across sessions. To avoid including such duplicates, we removed any bill that had a 96\% cosine similarity to an existing bill in the dataset. In addition, we ensured that the remaining bills with duplicate titles were all in the train partition. For additional details about this procedure, see Appendix \ref{sec:duplicates}.}

\begin{figure}[h]
    \centering

    \begin{subfigure}[t]{.5\linewidth}
         \centering
        \includegraphics[width=\linewidth]{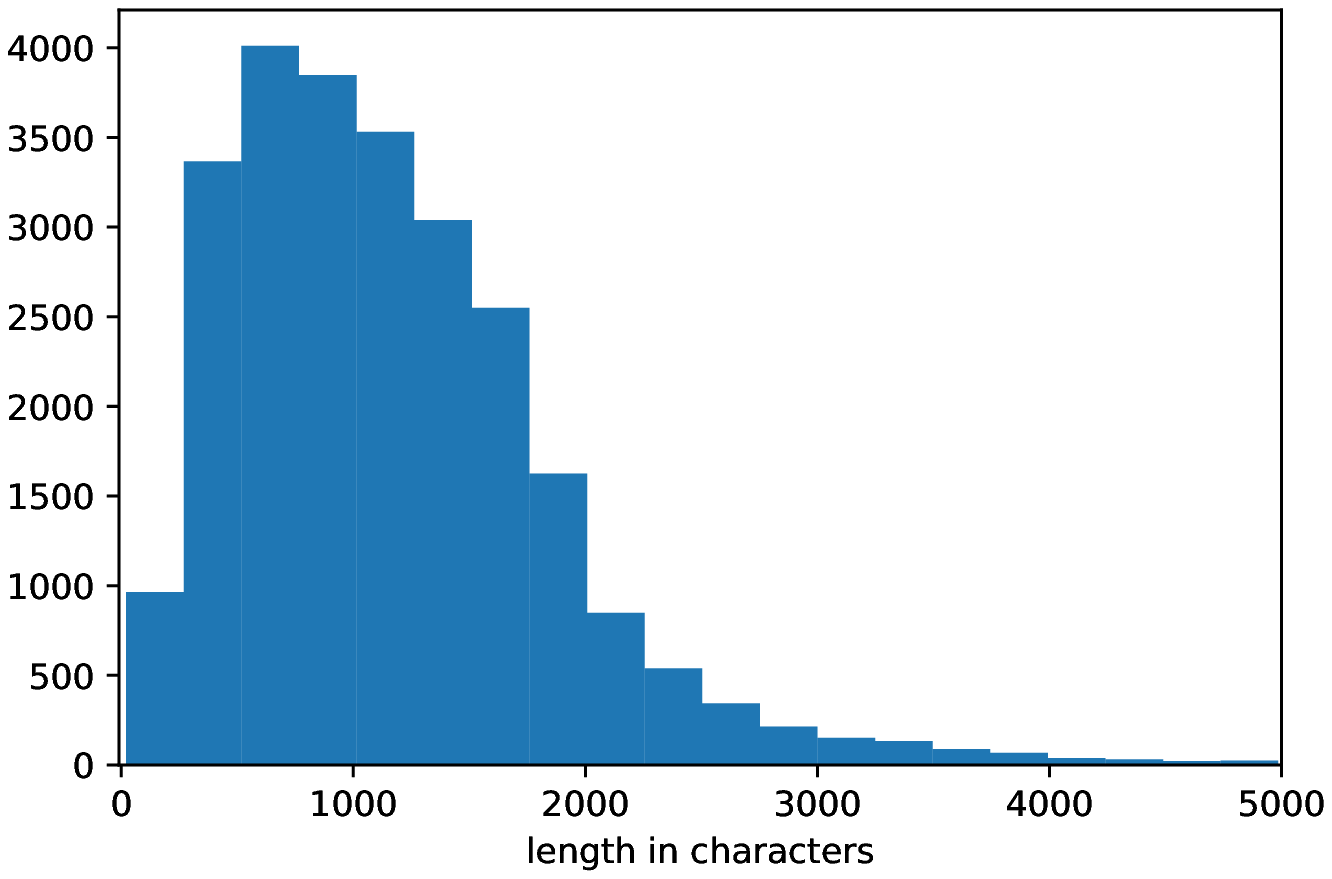}
         \caption{US Bill Summaries}
         \label{fig:conlen:summary}
     \end{subfigure}%
        ~
         \begin{subfigure}[t]{.5\linewidth}
         \centering
        \includegraphics[width=\linewidth]{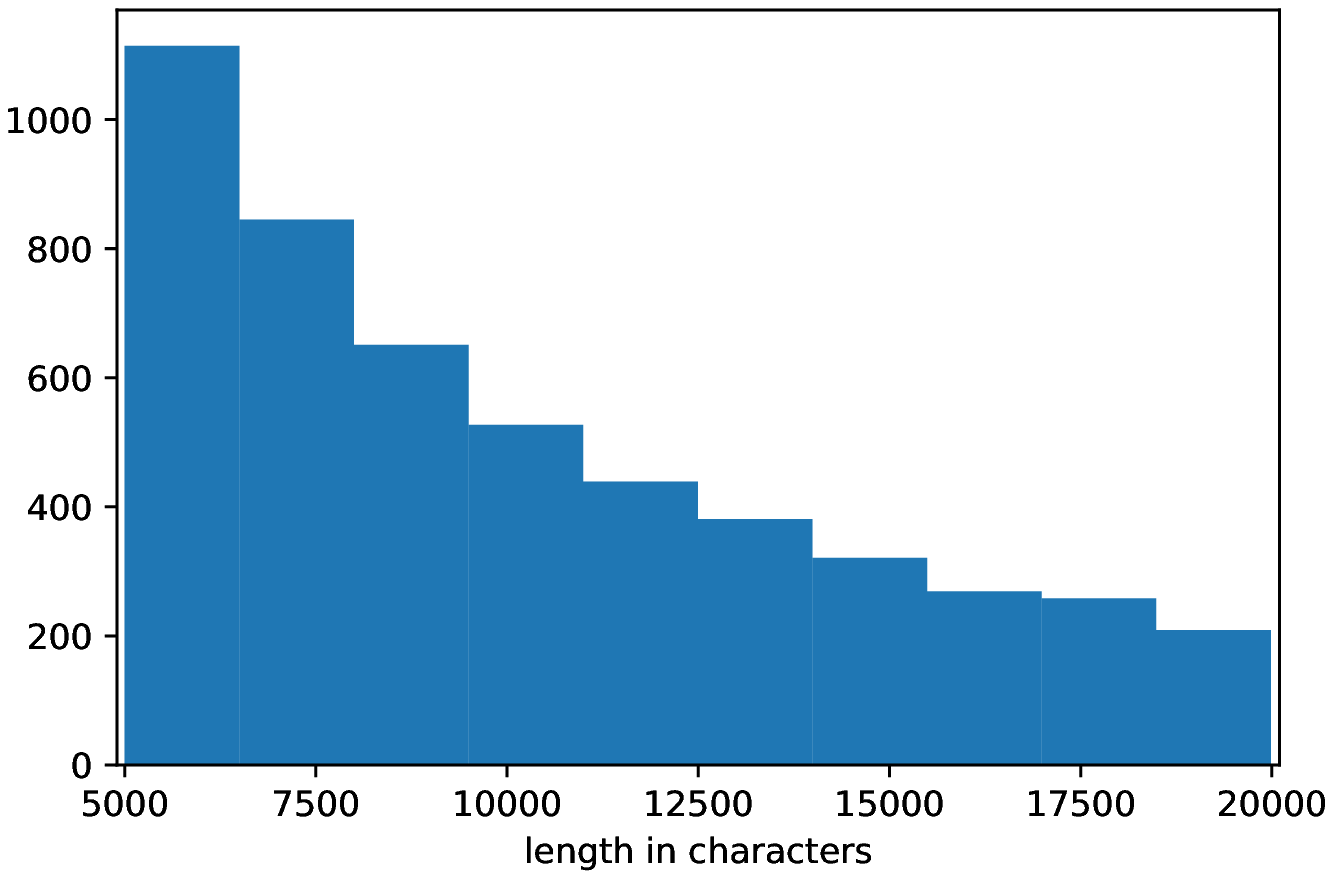}
         \caption{US Bill Text}
         \label{fig:conlen:text}
     \end{subfigure}
    
     \begin{subfigure}[t]{.5\linewidth}
         \centering
        \includegraphics[width=\linewidth]{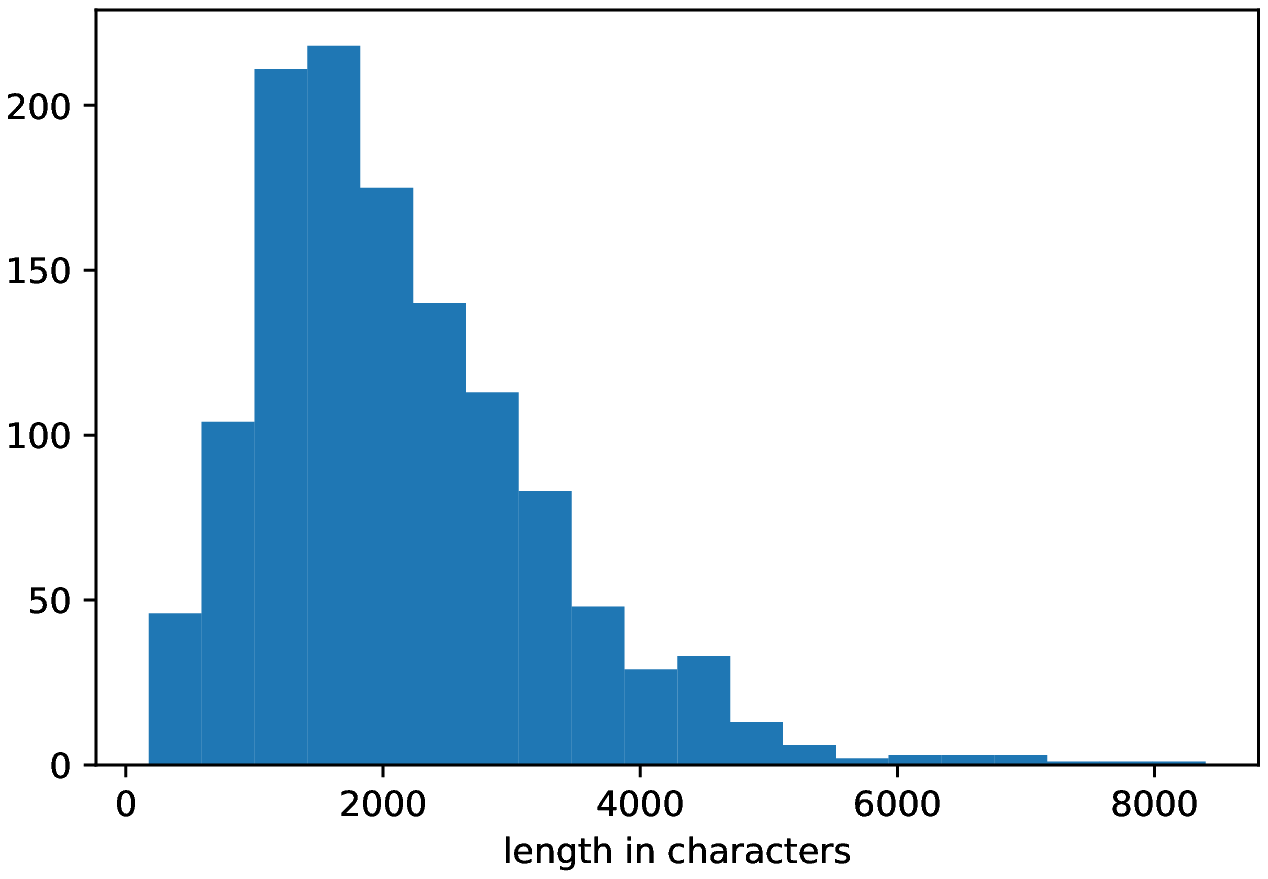}
         \caption{CA Bill Summaries}
         \label{fig:conlen:casum}
     \end{subfigure}%
        ~
    \begin{subfigure}[t]{.5\linewidth}
         \centering
        \includegraphics[width=\linewidth]{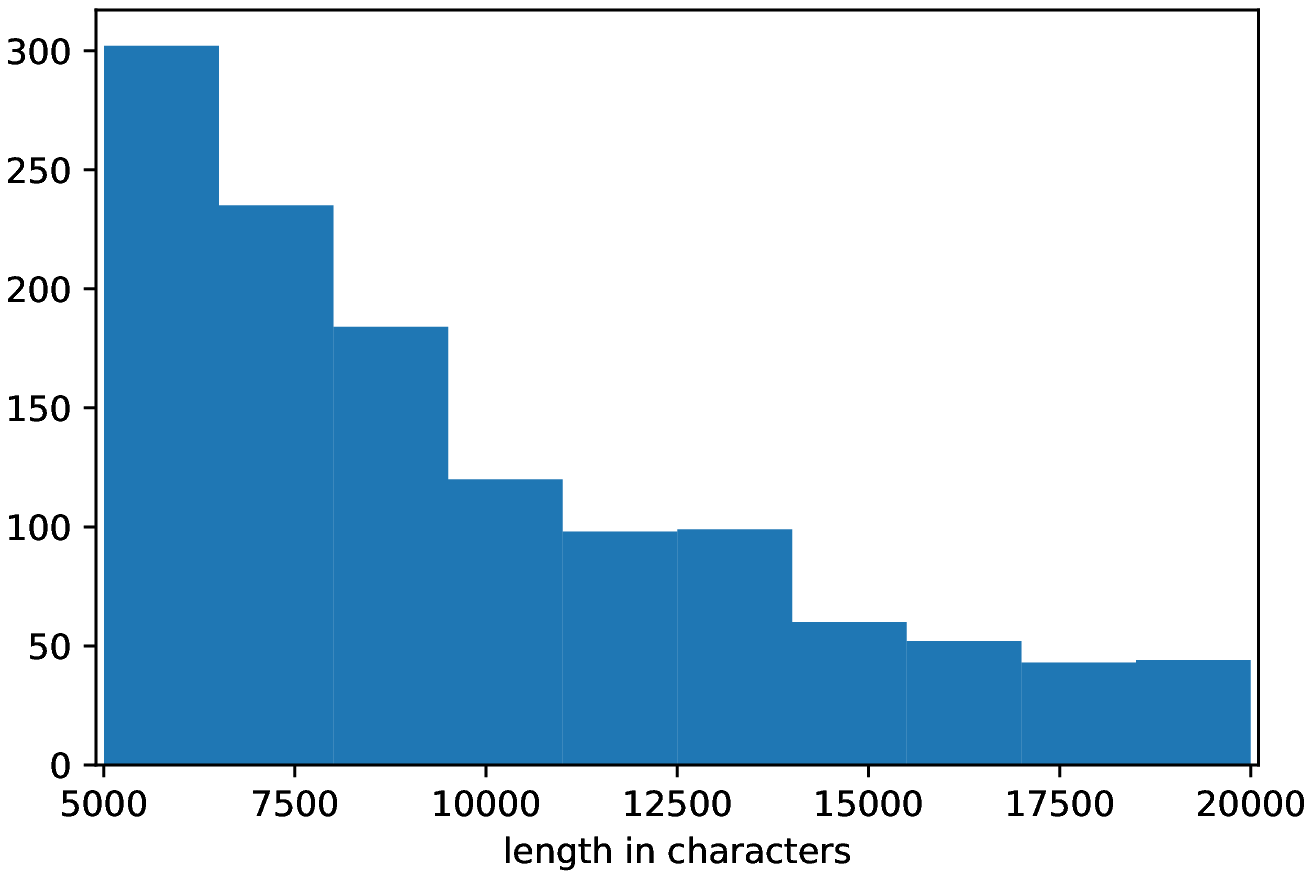}
         \caption{CA Bill Text}
         \label{fig:conlen:catext}
     \end{subfigure}
    
     \caption{Bill Lengths}
    \label{fig:conlen}

\end{figure}

For the summaries, we chose a 2000 character limit as 90\% of summaries are of this length or shorter; the limit here is, also, set in characters to be consistent with our document length cut-offs. The distribution of both text and summary lengths is shown in Figure ~\ref{fig:conlen}. Interestingly, there is little correlation between the bill and human summary length, with most summaries ranging from 1000 to 2000 characters.

For a closer comparison to other datasets, Table ~\ref{table:textlenclean} provides statistics on the number of words in the texts, after we simplify the structure of the texts.

\begin{table*}[]
\centering
\begin{tabular}{ll|l|l|l|l|l|l|}
\cline{3-8}
                                                 &    & mean & min & 25th & 50th & 75th & max  \\ \hline
\multicolumn{1}{|l|}{\multirow{2}{*}{Words}}     & US & 1382 & 245 & 923  & 1253 & 1758 & 8785 \\ \cline{2-8} 
\multicolumn{1}{|l|}{}                           & CA & 1684 & 561 & 1123 & 1498 & 2113 & 3795 \\ \hline
\multicolumn{1}{|l|}{\multirow{2}{*}{Sentences}} & US & 46   & 3   & 31   & 42   & 58   & 372  \\ \cline{2-8} 
\multicolumn{1}{|l|}{}                           & CA & 47   & 12  & 31   & 42   & 59   & 137  \\ \hline
\end{tabular}
 \caption{Text length distributions on preprocessed texts.}
 \label{table:textlenclean}
\end{table*}



\begin{figure}[h]
   \includegraphics[width=\linewidth]{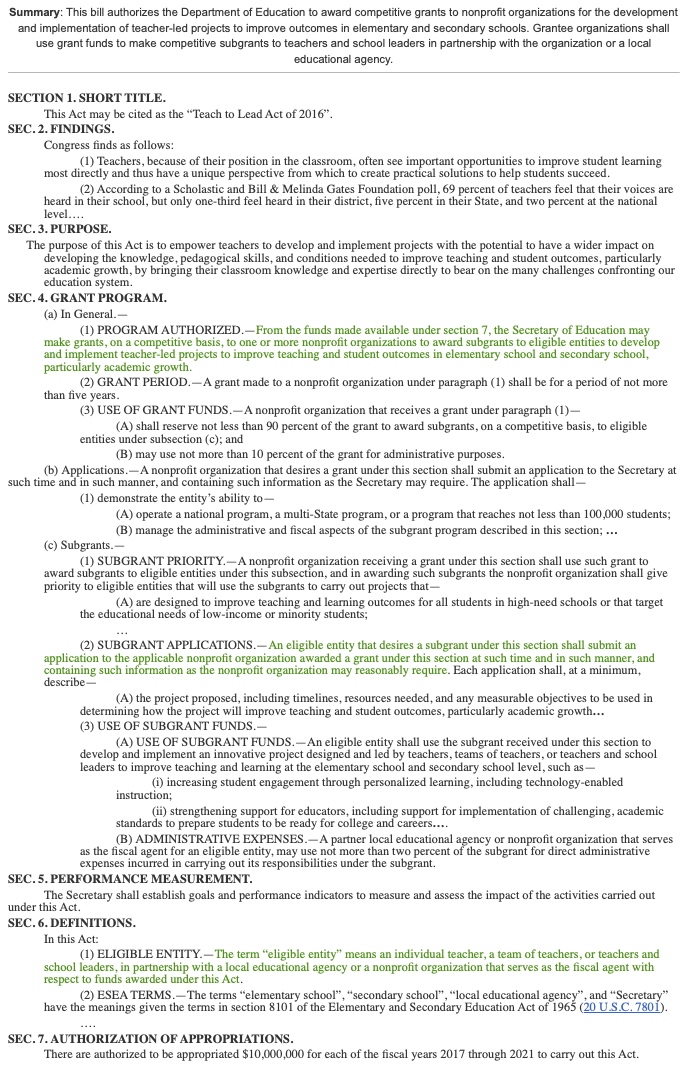}
    \caption{Example US Bill}
    \label{fig:trickybill}
    \vspace{-1.2mm}
\end{figure}

Stylistically, the BillSum dataset differs from other summarization corpora. Figure ~\ref{fig:trickybill} presents an example Congressional bill. The nested, bulleted structure is common to most bills, where each bullet can represent a sentence or a phrase. Yet, content-wise, this is a straightforward example that states key details about the proposed grant in the outer bullets. In more challenging cases, the bill may state edits to an existing law, without whose context the change is hard to interpret, such as:

\textit{Section 4 of the Endangered Species Act of 1973 (16 U.S.C. 1533) is amended in subsection (a)— in paragraph (1), by inserting ``with the consent of the Governor of each State in which the endangered species or threatened species is present"}\\

The average bill will contain both types of language, encouraging the study of both domain-specific and general summarization methods on this dataset. 

\section{Benchmark Methods}
\label{sec:methods}
To establish benchmarks on summarization performance, we evaluate several 
extractive summarization approaches by first scoring individual sentences, then using a selection strategy to pick the best subset \cite{Carbonell:1998:UMD:290941.291025}. While we briefly considered abstractive summarization models \cite{N16-1012}, we found that the existing models trained on news and Wikipedia data produced ungrammatical results, and that the size of dataset is insufficient for the necessary retraining. Recent works have successfully fine-tuned models for other NLP tasks to specific domains \cite{biobert}, but we leave to future work the exploration of similar abstractive strategies.

The scoring task is framed as a supervised learning problem. First, we create a binary label for each sentence indicating whether it belongs in the summary 
\cite{gillick2008icsi}.\footnote{As noted in Section \ref{sec:data}, it is difficult to define sentence boundaries for this task due to the bulleted structure of the documents. We simplify the text with the following heuristic: if a bullet is shorter than 10 words, we treat it as a part of the previous sentence; otherwise, we treat it as a full sentence. This cut-off was chosen by manually analyzing a sample of sentences. A more sophisticated strategy would be to check if each bullet is a sentence fragment with a syntactic parser and then reconstruct full sentences; however, the former approach is sufficient for most documents.} 
We compute a Rouge-2 Precision score of a sentence relative to the reference summary and simplify it to a binary value based on whether it is above or below $0.1$ \cite{lin2004rouge, zopfsentregression}. As an example, the sentences in the positive class are highlighted in green in Figure ~\ref{fig:trickybill}. 


Second, we build several models to predict the label. For the models, we consider two aspects of a sentence: its importance in the context of the document (\ref{sec:doc}) and its general summary-like properties (\ref{sec:sum}). 

\subsection{Document Context Model (DOC)}
\label{sec:doc}
A good summary sentence contains the main ideas mentioned in the document. Thus, researchers have designed a multitude of features to capture this property. We evaluate how several common ones transfer to our task:

The position of the sentence can determine how informative the sentence is \cite{seki2002tfidf}. 
We encode this feature as a fraction of `sentence position / total sentence count', to restrict this feature to the $0-1$ range regardless of the particular document's length. In addition, we include a binary feature for whether the sentence is near a section header.

An informative sentences will contain words that are important to a given document relative to others. Following a large percentage of previous works, we capture this property using TF-IDF \cite{seki2002tfidf, ramos2003tfidf}. First, we calculate a document-level TF-IDF weight for each word, then take the average and the maximum of these weights for a sentence as features. To relate language between sentences, ``sentence-level'' TF-IDF features are created using each sentence as a document for the background corpus; the average and max of the sentence's word weights are used as features.

We train a random forest ensemble model over these features with 50 estimators \cite{Breiman:randomforest}.\footnote{Implemented with \url{scikit-learn.org}} This method was chosen because it best captured the interactions between the small number of features.

\subsection{Summary Language Model (SUM)}
\label{sec:sum}
We hypothesize that certain language is more common in summaries than in bill texts. Specifically, that summaries primarily contain general effects of the bill (e.g awarding a grant) while language detailing the administrative changes will only appear in the text (e.g inserting or modifying relatively minor language to an existing statute). Thus, a good summary should contain only  the major actions.

\citet{hong2014improving} quantify this aspect using hand-engineered features based on the the likelihood of words appearing in summaries as opposed to the text. Later, \citet{cao2015learning} built a Convolutional Neural Network (CNN) to predict if a sentence belongs in the summary and showed that this straightforward network outperforms 
engineered features. We follow their approach, using the BERT model as our classifier \cite{devlin2018bert}. BERT can be adapted for and has achieved state-of-the-art performance on a number of NLP tasks, including binary sentiment classification.\footnote{All code described are used directly from \url{https://github.com/google-research/bert}}

To adapt the model to our domain, we pre-train the \textbf{Bert-Large Uncased} model on the ``next-sentence prediction'' task using the US training dataset for 20,000 steps with a batch size of 32.\footnote{This is the pretraining procedure recommended by the authors of BERT on their github website.} The pretraining stategy has been successfully applied to tune BERT for tasks in the biomedical domain \cite{biobert}. Using the pretrained model, the classification setup for BERT is trained on sentences and binary labels for 3 epochs over the training data.

\subsection{Ensemble and Sentence Selection}
To combine the signals from the DOC and SUM models, we create an ensemble averaging the two probability outputs.\footnote{Additional experiments using Linear Regression with the actual Rouge-2 Precision score as the target, but found that they produced similar results.}

To create the final summary, we apply the Maximal Marginal Relevance (MMR) algorithm \cite{goldstein2000mmr}. MMR iteratively constructs a summary by including the highest scoring sentence with the following formula:

\[s_{next} = \max\limits_{s \in D - S_{cur}} 0.7 * f(s) - 0.3 * \textit{sim}(s, S_{cur}) \]

where $D$ is the set of all the sentences in the document, $S_{cur}$ are the sentences in the summary so far, $f(s)$ is the sentence score from the model, $\textit{sim}$ is the cosine similarity of the sentence to $S_{cur}$, and $0.7$ and $0.3$ are constants chosen experimentally to balance the two properties. This method allows us to pick relevant sentences while minimizing redundancies. We repeat this process until we reach the length limit of 2000 characters.

\section{Results}

To estimate the upper bound on our approach, an oracle summarizer is created by using the true Rouge-2 Precision scores with the MMR selection strategy. In addition, we evaluate the following unsupervised baselines: \textbf{SumBasic} \cite{nenkova2005sumbasic}, Latent Semantic Analysis (LSA) \cite{gong2001lsa} and \textbf{TextRank} \cite{mihalcea2004textrank}.  The final results are shown in Table~\ref{tab:res}. The Rouge F-Score is used because it considers both the completeness and conciseness of the summary method.\footnote{Precision and recall scores are listed in the supplemental material for additional context.}$^,$\footnote{Rouge scores calculated using \url{https://github.com/pcyin/PyRouge}}

We evaluated the DOC, SUM, and ensemble classifiers separately. All three of our models outperform the other baselines, demonstrating that there is a ``summary-like'' signal in the language across bills. The SUM model outperforms the DOC model showing that a strong language model can capture general summary-like features; this result is in line with \citet{cao2015learning} and \citet{collins2017supervised} sentence level neural network performance. However, in those studies incorporating several contextual features improved the performance, while DOC+SUM performs similarly to DOC. In future work we plan to incorporate contextual features into the neural network directly; \citet{collins2017supervised} showed that this strategy is effective for scientific article summarization. In addition, we plan to explore additional sentence selection strategies instead of always adding sentences up to the 2000 character limit.

Next, we applied our US model to CA bills. Overall, the performance is 
lower than on US bills (Table~\ref{tab:res:ca}), but all three supervised methods perform better than the unsupervised baselines, suggesting that models built using the language of US Bills can transfer to other states. Interestingly, the SUM model performs similarly to the DOC in the CA dataset, suggesting that the BERT model may have overfit to the US language. An additional reason for the similar performance is the difference in the structure of the summaries: In California the provided summaries state not only the proposed changes, but the relevant pieces of the existing law, as well (see Appendix ~\ref{appendix:cali} for a more in-depth discussion). We hypothesize that a model trained on multi-state data would transfer better, thus we plan to expand the dataset to include all twenty-three states with human-written summaries.

\begin{table}[h]
  \caption{ROUGE F-scores (\%) of different methods.}\label{tab:res}
    \begin{tabular}{|l|c|c|c|}
    \hline
     &  Rouge-1 & Rouge-2 & Rouge-L \\ \hline
     Oracle & 45.11  & 28.74 & 37.38 \\
    \hline
    SumBasic &  30.74 & 14.16 & 23.92\\ 
    LSA &  32.64 & 15.69 & 26.26\\ 
    TextRank &  34.35 & 17.77 & 27.80\\ 
    \hline 
    DOC & 38.51 & 21.38 & 31.49 \\
    SUM &  40.69 & 23.88 & 33.65\\ 
    DOC + SUM &  40.80 & 23.83 & 33.73\\ 

    \hline
    \end{tabular}\\
  \subcaption{Congressional Bills}\label{tab:res:bill}
    \begin{tabular}{|l|c|c|c|}
    \hline
     &  Rouge-1 & Rouge-2 & Rouge-L \\ \hline
    Oracle & 48.61 &  32.83 &  41.94 \\
    \hline
    SumBasic &  35.47 & 16.18 & 29.98\\ 
    LSA &  35.06 & 16.34 & 29.93\\ 
    TextRank &  35.81 & 18.10 & 29.97\\ 
    \hline 
    DOC &  38.35 & 19.76 & 32.80 \\ 
    SUM &  38.90 & 20.79  & 33.20 \\ 
    DOC + SUM &  39.65 & 21.14 & 34.05\\ 

    \hline
   
    \end{tabular}\\
    \subcaption{CA Bills}\label{tab:res:ca}
    \vspace{-7mm}
\end{table}

\subsection{Summary Language Analysis}
\label{sec:methods:sum}
The success of the SUM model suggests that certain language is more summary-like. Following a study by \citet{hong2014improving} on news summarization, we apply KL-divergence based metrics to quantify which words were more summary-like. The metrics are calculated by:

\begin{enumerate}

    \item Calculate the probability of unigrams appearing in the bill text and in the summaries ($P_t(w)$ and $P_s(w)$ respectively). 
    \item Calculate KL scores as : $KL_w(S | T) = P_s(w) * \ln{\frac{P_s(w)}{P_t(w)}}$ and the opposite. 
\end{enumerate}

A large value of $KL(S|T)$ indicates that the word is summary-like and $KL(T|S)$ indicates a text-like word. Table \ref{tab:kl_examples} shows the most summary-like and text-like words in bills and resolutions. For both document types, the summary-like words tend to be verbs or department names; the text-like words mostly refer to types of edits or background content (e.g ``reporting the rise of..''). This follows our intuition about summaries being more action driven. While a complex model, like BERT, may capture these signals internally; understanding the significant language explicitly is important both for interpret ability and for guiding future models.

\begin{table}[h]
\caption{Examples of summary and text like words}
\label{tab:kl_examples}
\begin{tabular}{|l|p{1.72in}|}
\hline
Summary-like & prohibit, DOD, VA, allow, penalty, prohibit, EPA, eliminate, implement, require \\ \hline
Text-like & estimate, average, report, rise, section, finish, percent, debate
 \\ \hline

\end{tabular}
\end{table}

\section{Conclusion}
In this paper, we introduced BillSum, the first corpus for legislative summarization. This is a challenging summarization dataset due to the technical nature and complex structure of the bills. We have established several baselines and demonstrated that there is a large gap in performance relative to the oracle, showing that the problem has ample room for further development. We have also shown that summarization methods trained on US Bills transfer to California bills - thus, the summarization methods developed on this dataset could be used for legislatures without human written summaries.

\bibliography{emnlp-billsum}

\bibliographystyle{acl_natbib_ref}

\appendix

\section{Duplicate Removal Procedure}
\label{sec:duplicates}
There are a number of reasons near duplicate bills are written by Congress, including the introduction of companion bills in the House and Senate during the same session, and the reintroduction across sessions of bills that failed to pass during a previous session.
To avoid including duplicate examples in the dataset 
we looked for similar bills by: 
\begin{enumerate}
    \item Vectorizing the text using \texttt{scikit-learn}'s CountVectorizer.
    \item Removing the top $15\%$ of most common words and generic stop words.
    \item Computing cosine similarity between the texts and the summaries for each pair of bills and averaging the two similarities.
    \item Iteratively adding bills to the dataset, skipping examples that were more than $96\%$ similar to any bills already added.
\end{enumerate} 

After this procedure is run, the data still includes some bills with identical titles. This can happen for two reasons: either the title is generic and refers to two unrelated bills, or one is a reintroduction of the other with enough modified content to not be considered a duplicate. We put all the bills with identical titles in the train partition.

\section{Additional ROUGE Scores}

As discussed in the Results section, F-Scores encourage a balance between comprehensiveness and conciseness. However, as it is useful to analyze the precision and recall scores separately, both are presented in Table~\ref{tab:rouge} for US Bills and in Table~\ref{tab:rougeca} for CA Bills. All tested methods favor recall, since they consistently generate a 2000 character summary, instead of stopping early when a concise summary may be sufficient. For both datasets, the difference in Recall between the Oracle and DOC+SUM summarizer is a lot smaller than for Recall; which suggests that a lot of useful summary content can be found with an extractive method. In future work, we will focus on extracting more granular snippets to improve precision.

\begin{table}[h]
\caption{ROUGE Scores of Congressional Bills}\label{tab:rouge}
    \begin{tabular}{|l|c|c|c|}
    \hline
         &  Rouge-1 & Rouge-2 & Rouge-L \\ \hline
         Oracle & 40.94  & 25.82 & 38.29 \\
        \hline
        SumBasic &  24.63 &	11.71 &	22.36\\ 
        LSA &  27.34 &	13.30 &	24.96\\ 
        TextRank &  29.86 &	15.37 &	26.99\\ 
        \hline 
        DOC & 32.61 & 17.93  & 30.10 \\
        SUM &  34.59 & 20.15 & 32.18\\ 
        DOC + SUM &  34.77 & 20.11 & 32.21\\ 
    
        \hline
    \end{tabular}\\
    \subcaption{Precision Scores}
    \bigskip
    \begin{tabular}{|l|c|c|c|}
        \hline
         &  Rouge-1 & Rouge-2 & Rouge-L \\ \hline
         Oracle & 58.19  & 39.17 & 54.52 \\
        \hline
        SumBasic &  47.37 &	21.90 &	42.88\\ 
        LSA &  46.53 & 23.10 &	42.32 \\ 
        TextRank &  46.49	& 25.36 &	41.88 \\ 
        \hline 
        DOC & 54.16 & 32.31 & 49.92 \\
        SUM &  56.68 & 35.56 & 52.62 \\ 
        DOC + SUM &  56.69 & 35.56 & 52.62\\ 
        \hline
    \end{tabular}\\
        \subcaption{Recall Scores}
\end{table}

\begin{table}[!htbp]
\caption{ROUGE Scores of California Bills}\label{tab:rougeca}
    \begin{tabular}{|l|c|c|c|}
    \hline
         &  Rouge-1 & Rouge-2 & Rouge-L \\ \hline
         Oracle & 45.00  & 30.85 & 42.79 \\
        \hline
        SumBasic &  33.72 &	16.30 &	30.41\\ 
        LSA &  34.84 &	17.24 &	31.69\\ 
        TextRank &  36.66 &	19.28 &	32.61\\ 
        \hline 
        DOC & 38.31 & 20.49  & 34.67 \\
        SUM &  41.67 & 22.10 & 37.45\\ 
        DOC + SUM &  39.86 & 22.34 & 36.17\\ 
    
        \hline
    \end{tabular}\\
    \subcaption{Precision Scores}
    \begin{tabular}{|l|c|c|c|}
        \hline
         &  Rouge-1 & Rouge-2 & Rouge-L \\ \hline
         Oracle & 62.96  & 43.83 & 59.58 \\
        \hline
        SumBasic &  40.47 &	17.89 &	36.36\\ 
        LSA &  38.23 &	17.28 &	34.61 \\ 
        TextRank &  37.79 &	18.97 &	33.50 \\ 
        \hline 
        DOC & 41.50 & 21.33 &  37.42 \\
        SUM &  39.04 & 21.73 & 35.25 \\ 
        DOC + SUM &  42.51 & 22.82 & 38.41 \\ 
        \hline
    \end{tabular}\\
        \subcaption{Recall Scores}
    \end{table}

\section{Additional Bill Examples}

We highlight several example bills to showcase the different types of bills found in the dataset.

\subsection{Complex Structure Example}

In the Data section, we discussed some of the challenges with processing bills: complex formatting and technical language. Figure~\ref{fig:worstbill} is an excerpt from a particularly difficult example:  

\begin{figure}[!h]
   \includegraphics[width=\linewidth]{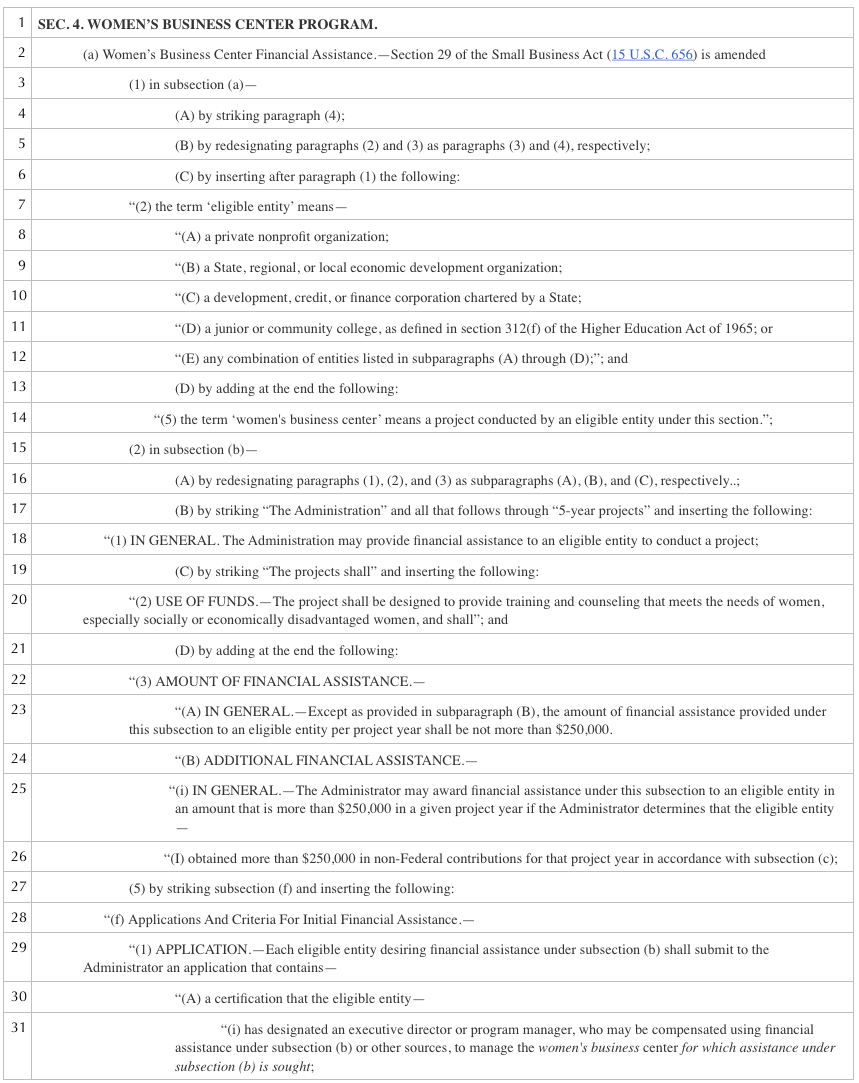}
    \caption{US H.R.1680 (115th)}
    \label{fig:worstbill}
    \vspace{-1.2mm}
\end{figure}

\begin{itemize}
    \item The text interleaves several layers of bullets. Lines 3, 15, 27 represent the same level (points (3) and (4) omitted for space); lines 16, 17, 19 and 21 go together, as well. These multiple levels need to be handled carefully, or the summarizer will extract snippets that can not be interpreted without context.
    \item Lines 22-26 both introduce new language for the law and use the bulleted structure.
    \item Line 27 states that the existing ``subsection (f)'' is being removed and replaced. While lines 28 onward state the new text, the meaning of the change relative to the current text is not clear.
\end{itemize}

The human-written summary for this bill was:\\

\begin{it}
(Sec. 4)``Women's business center" shall mean a project conducted by any of the following eligible entities:
\begin{itemize}
    \item a private nonprofit organization;
    \item a state, regional, or local economic development organization;
    \item a state-chartered development, credit, or finance corporation;
    \item a junior or community college; or
    \item any combination of these entities.
\end{itemize}
The SBA may award up to \$250,000 of financial assistance to eligible entities per project year to conduct projects designed to provide training and counseling meeting the needs of women, especially socially and economically disadvantaged women.
\end{it}\\

Most of the relevant details are capture in the text between lines 8-14 and 20-24. For examples similar to this one, the summary language is extracted almost directly from the text, but, parsing them correctly from the original structure is a non-trivial task.

\subsection{Paraphrase Example}
\begin{figure}[!h]
    \includegraphics[width=\linewidth]{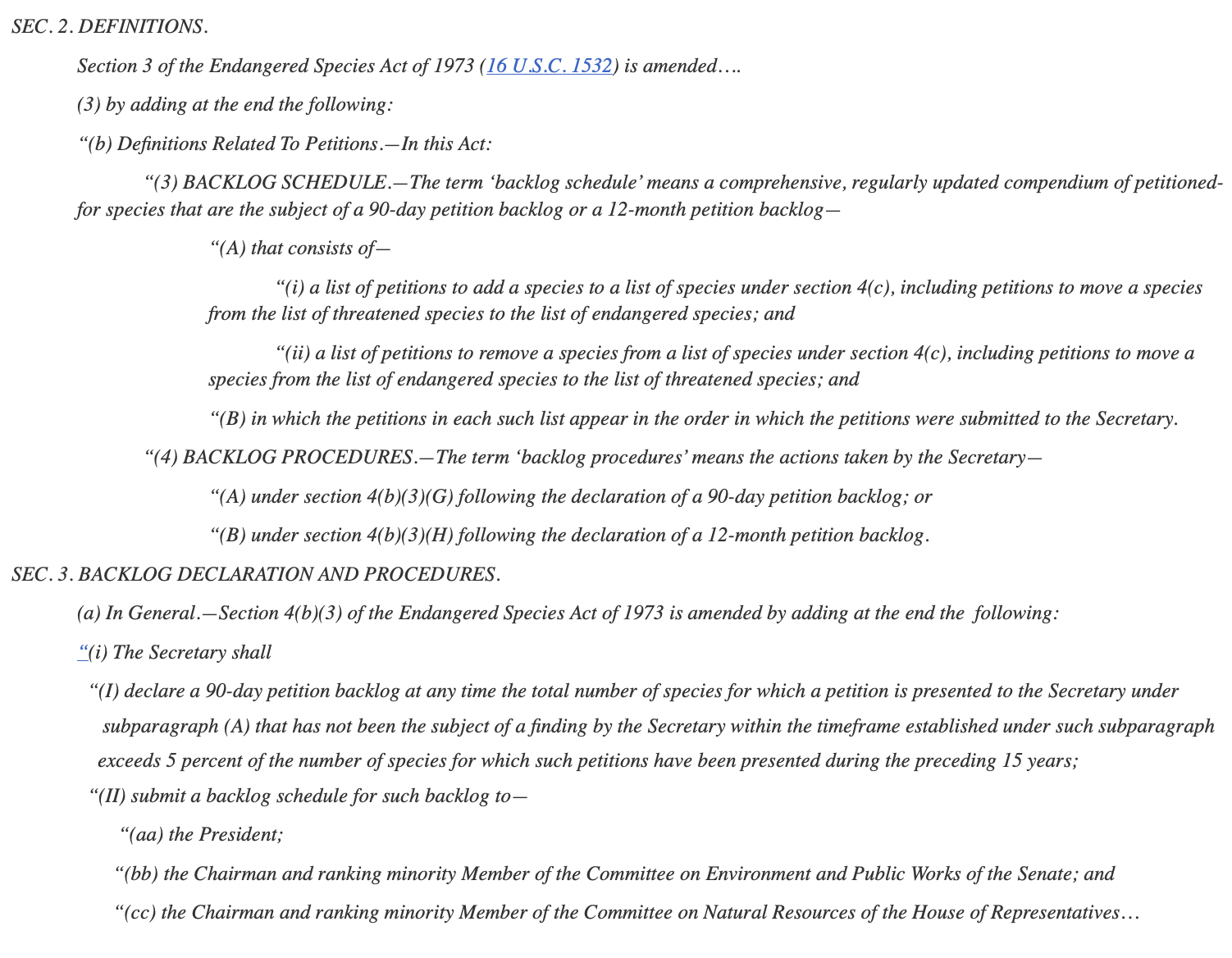}
    \caption{US H.R.6355 (115th)}
    \label{fig:paraphrase}
\end{figure}

For a subset of the bills, the CRS will paraphrase the technical language. In these cases, extractive summarization methods are particularly limited. Consider the example in Figure~\ref{fig:paraphrase} and its summary:

\textit{This bill amends the Endangered Species Act of 1973 to revise the process by which the Department of the Interior or the Department of Commerce, as appropriate, reviews petitions to list a species on the endangered or threatened species list. Specifically, the bill establishes a process for the appropriate department to declare a petition backlog and discharge the petitions when there is a backlog.}\\

While the bill elaborates of the ```process'', the summary states that one was created. This type of summary would be hard to construct by a purely extractive method. 

\subsection{California Example}
\label{appendix:cali}
\begin{figure}[!h]
    \includegraphics[width=\linewidth]{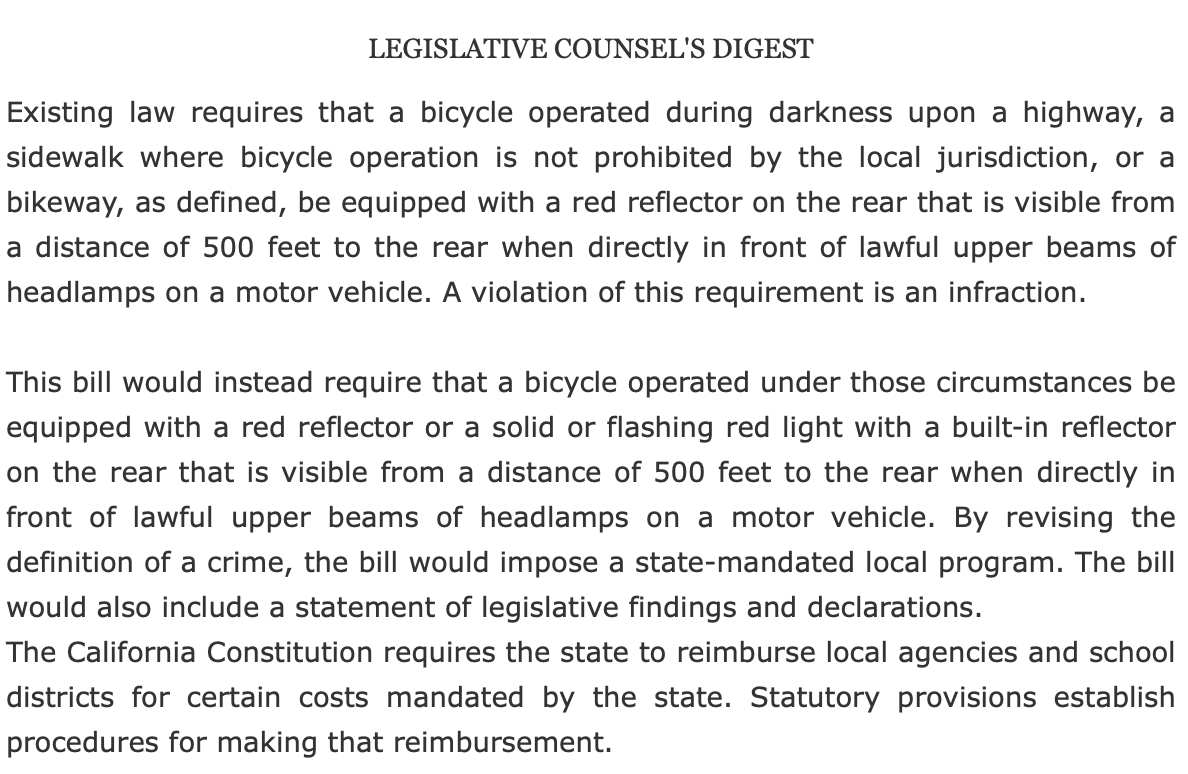}
    \caption{California Bill Summary}
    \label{fig:cabill}
\end{figure}

The California bills follow the same general patterns as US bills, but the format of some summaries is different. In Figure ~\ref{fig:cabill}: the summary, first, explains the existing law, then explains the change. The additional context is useful, and in the future we may build a system that references the existing law to create better summaries. 

\end{document}